\documentclass{article}
\usepackage{spconf,amsmath,graphicx}
\usepackage{bbding}
\usepackage{threeparttable}

\title{Synthetic Pseudo Anomalies for Unsupervised Video Anomaly Detection: A Simple yet Efficient Framework based on Masked Autoencoder}

\name{Xiangyu Huang$^{1}$, Caidan Zhao$^{1*}$ \thanks{*Corresponding Author. This work was supported in part by the National Natural Science Foundation of China under Grant No. 61971368, No. U20A20162 and No. 61731012, and in part by the Natural Science Foundation of Fujian Province of China No. 2019J01003.}, Chenxing Gao$^{1}$, Lvdong Chen$^{1}$, and Zhiqiang Wu$^{2}$}

\address{$^{1}$ School of Informatics, Xiamen University \\
$^{2}$ PKU-Wuhan Institute for Artificial Intelligence}

\begin{document}
%
\maketitle
\begin{abstract}
Due to the limited availability of anomalous samples for training, video anomaly detection is commonly viewed as a one-class classification problem. Many prevalent methods investigate the reconstruction difference produced by AutoEncoders (AEs) under the assumption that the AEs would reconstruct the normal data well while reconstructing anomalies poorly. However, even with only normal data training, the AEs often reconstruct anomalies well, which depletes their anomaly detection performance. To alleviate this issue, we propose a simple yet efficient framework for video anomaly detection. The pseudo anomaly samples are introduced, which are synthesized from only normal data by embedding random mask tokens without extra data processing. We also propose a normalcy consistency training strategy that encourages the AEs to better learn the regular knowledge from normal and corresponding pseudo anomaly data. This way, the AEs learn more distinct reconstruction boundaries between normal and abnormal data, resulting in superior anomaly discrimination capability. Experimental results demonstrate the effectiveness of the proposed method.
\end{abstract}

\section{Introduction}
\label{sec:intro}

Video anomaly detection (VAD) refers to identifying events that do not conform to expected behavior \cite{chandola2009anomaly} in surveillance videos. This task is extremely challenging due to the limited availability of anomalous samples: anomalous events rarely occur in the real world, and the forms of abnormal events are unexpected. It is therefore not feasible to collect sufficient anomaly examples for training a fully-supervised binary classification model. Consequently, VAD is typically seen as a one-class classification (OCC) problem in which only normal examples are used to train a novelty detection model \cite{liu2018future, antic2011video, hasan2016learning, gong2019memorizing, park2020learning, shen2022video}. Then the events that deviate from the regular learned representations are regarded as anomalies. \\
\indent
Prevalent VAD methods follow a reconstruction paradigm. Specifically, they train the AEs to extract the feature representations of normal data by minimizing reconstruction errors. Then the trained model is expected to reconstruct the abnormal data with larger reconstruction errors at test time, making abnormal data detectable from normal ones. However, several researchers \cite{gong2019memorizing, park2020learning, munawar2017limiting} observe that AEs sometimes reconstruct anomalies well, indicating that the reconstruction difference between normal and abnormal data may not be discriminative enough to detect the anomalies. \\
\indent
Some approaches can alleviate the aforementioned limitations by using pseudo-anomaly samples synthesized from only the normal data  \cite{zaheer2020old, astrid2021synthetic, astrid2021learning}. For example, Astrid \textit{et al.} \cite{astrid2021learning} generate pseudo anomalies by various data augmentation techniques among image classification tasks. By simulating out-of-normal data distribution, such methods help AEs learn a vivid reconstruction boundary under the OCC setting. However, these methods need an extra phase to train a pseudo anomaly synthesizer or rely on a lot of well-designed data processing, which leads to unstable VAD performance and excessive training time. \\
\indent
Within the context of pseudo anomaly based methods, this paper proposes a simple yet efficient framework based on masked autoencoder. Compared to previous related works, we try to utilize random masked patches to generate pseudo anomaly samples. Inspired by the popular masked image modeling \cite{he2021masked, xie2021simmim}, we build a universal masked autoencoder architecture for VAD by embedding random mask tokens to simulate anomalies, which is a simple yet efficient synthetic method that avoids extra data processing in the original normal data. In addition, previous works simulate the real scene that anomalous events rarely occur and feed pseudo anomaly samples with a small probability in the training phase. This way, the pseudo anomaly samples are underutilized and have a trivial effect on the AEs. So we introduce a normalcy consistency training strategy for fully using the pseudo anomaly samples. Specifically, We minimize the bidirectional KL-divergence between the encoding features of the normal sample and the corresponding pseudo-anomaly sample. Our framework can build more distinct reconstruction boundaries between normal and abnormal data by learning consistent normalcy knowledge. The proposed framework demonstrates superior performance through experiments on VAD benchmarks.

\section{Methods}
\label{sec:format}

Given an original video frame $X^{O}$, we first utilize random masks to generate the corresponding pseudo anomaly video frame $X^{P}$. An encoder extracts the feature representations $f^{O}$ and $f^{P}$, respectively. Then, the extracted latent features $f^{O}$ and $f^{P}$ are used to predict the next frames $Y^{O}$ and $Y^{P}$ by a one-layer linear decoder, respectively. To make $Y^{O}$ and $Y^{P}$ close to their ground truth $Y$, we minimize their distance regarding intensity as well as the gradient. In addition, we minimize the KL-divergence between $f^{O}$ and $f^{P}$ for the purpose of encouraging the model to mine the consistent normalcy representations from normal samples and corresponding pseudo anomaly samples. Finally, the prediction error between the predicted frame and its ground truth determines whether it is a normal or abnormal frame. An overview of our proposed framework is illustrated in Figure 1. In the following subsections, we will present all the components of our framework in detail.

\noindent
\textbf{Frame Prediction for Video Anomaly Detection. }
The future frame prediction paradigm is commonly used for Video Anomaly Detection. Existing works \cite{liu2018future,park2020learning,cai2021appearance} often use the designed autoencoder (AE) to tackle this problem: an encoder learns to extract features from only normal training video frames, and a decoder generates the predicted target frame by using the extracted features. For an input frame X, the above process can be defined as:
\begin{equation}
\hat{Y}=\mathcal{D}(\mathcal{E}(X))
\end{equation}
where $\mathcal{E}$ and $\mathcal{D}$ are encoder and decoder, respectively. And the training target is to make the predicted frame $\hat{Y}$ close to its ground truth $Y$ while the anomalies will generate the larger prediction error to be spotted at test time.

\noindent
\textbf{Pseudo Anomaly Strategy based on Random Masking. }
Due to the unavailability of anomalies during training, the AE-based methods often cannot discriminate anomalies from normal data at test time. Inspired by \cite{astrid2021synthetic, astrid2021learning}, we introduce pseudo anomaly samples during training. Following the one-class classification problem setting, we do not use real abnormal samples, so pseudo anomaly samples are generated by altering normal data.
\begin{figure}[!htbp]
\centering
\includegraphics[width=85mm]{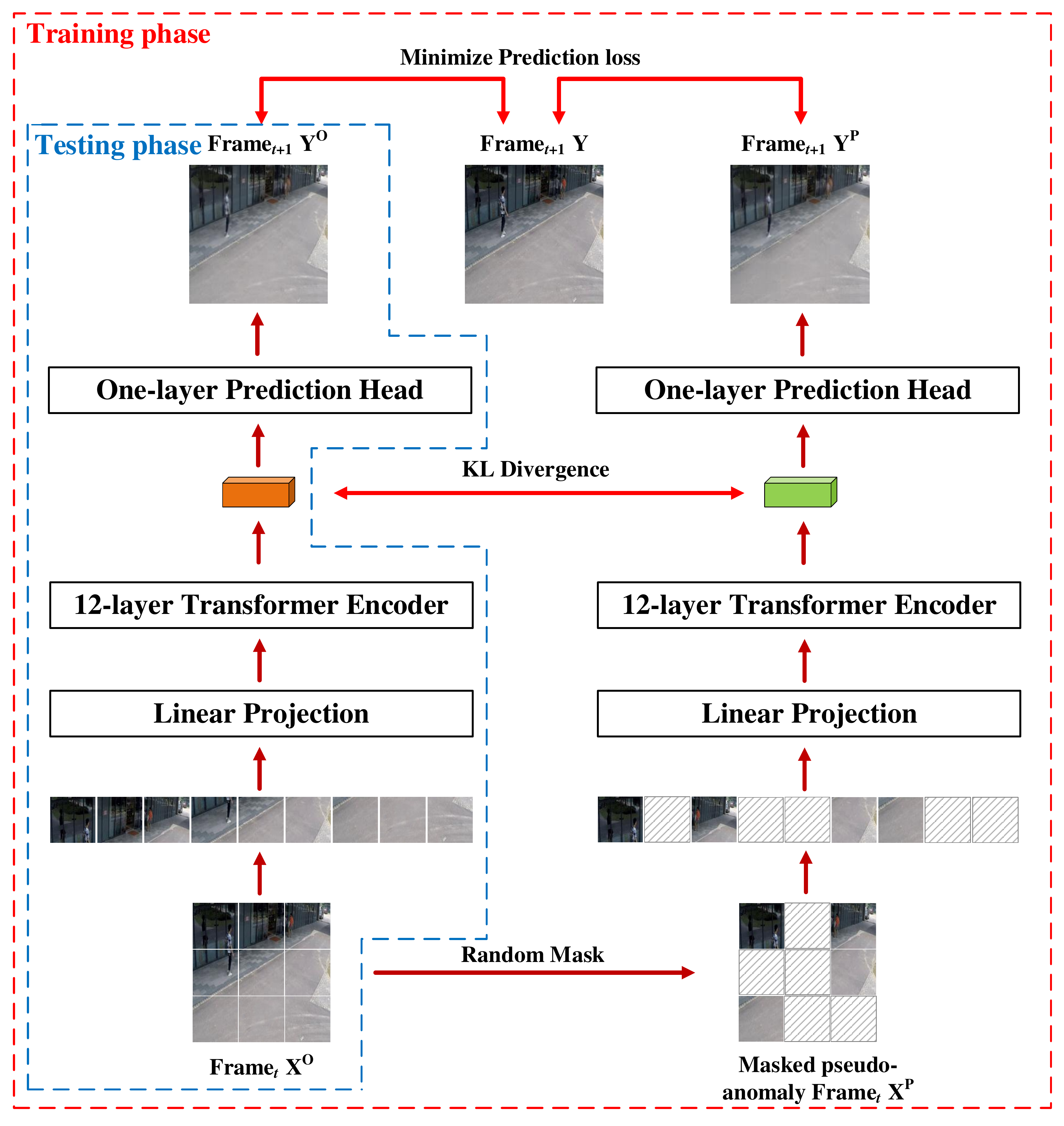}
\caption{Summary of the proposed framework.
}
\label{fig:overview}
\end{figure}
Unlike previous pseudo anomaly based approaches, we propose to use the patch-aligned random masking method to generate the pseudo anomaly samples. Image patches are the basic processing units of the vision Transformer, and it is convenient to operate the patch-level masking method. Therefore, for the architecture design of AE, we refer to \cite{xie2021simmim}, using a ViT-B encoder and a one-layer linear decoder. Compared with previous pseudo-anomaly based methods, our approach can also generate various pseudo anomaly samples whose pixel information differs from normal samples. Moreover, our approach is simple and efficient without complicated data augmentation methods.

\noindent
\textbf{Training. }
Following existing works based on future frame prediction paradigm \cite{liu2018future}, we use intensity and gradient loss to train original normal video frames $X^{O}$. We minimize intensity and gradient difference to make the normal prediction $Y^{O}$ close to its ground truth $Y$. Specifically, the intensity is defined as minimizing the $\ell_{2}$ distance between the predicted frame $Y^{O}$ and its ground truth $Y$ as follows:

\begin{equation}
	L_{i n t}(Y^{O},Y)=\left\|Y^{O}-Y\right\|_{2}^{2}
\end{equation}

The gradient loss is defined as follows:

\begin{equation}
	\begin{aligned}
		L_{g d}(Y^{O},Y)=\sum_{i, j}&\left\|\left|Y^{O}_{i, j}-Y^{O}_{i-1, j}\right|-\left|Y_{i, j}-Y_{i-1, j}\right|\right\|_{1}\\
		+&\left\|\left|Y^{O}_{i, j}-Y^{O}_{i, j-1}\right|-\left|Y_{i, j}-Y_{i, j-1}\right|\right\|_{1}
	\end{aligned}
\end{equation}
where $i$, $j$ denote the spatial index of a video frame.

So the loss function of normal video frames is defined as follows:
\begin{equation}
	L_{N}=L_{i n t}(Y^{O},Y) + L_{g d}(Y^{O},Y)
\end{equation}

For the generated pseudo anomaly video frames $X^{P}$, we encourage the model to predict only normal behavior in spite of abnormal input. Therefore, even if the input data contains abnormal regions, the model also learns to predict the normal regions well. It helps the model to learn vivid reconstruction boundaries between normal and abnormal data. Moreover, for the abnormal regions, the model may tend to predict normalcy representations, which will lead to a larger prediction error. So we also minimize intensity and gradient difference to make the pseudo anomaly prediction $Y^{P}$ close to its ground truth $Y$. The loss of pseudo anomaly video frames can be similarly defined as:

\begin{equation}
	L_{P}=L_{i n t}(Y^{P},Y) + L_{g d}(Y^{P},Y)
\end{equation}

Then, we propose a training strategy that differs from previous pseudo anomaly based works. Instead of inputting pseudo anomaly examples with a small probability, we simultaneously input both normal and corresponding pseudo-anomaly samples and mine the consistent normalcy representations between them. Specifically, we minimize the bidirectional KL-divergence between their encoding features $f^{O}$ and $f^{P}$. The consistency loss is defined as follows:

\begin{equation}
    L_{cst}=\frac{1}{2}\left(K L\left(f^{O} \| f^{P}\right)+K L\left(f^{P} \| f^{O}\right)\right)
\end{equation}

Finally, the overall loss takes the form as follows:
\begin{equation}
	L=\lambda_{N} L_{N}+\lambda_{P} L_{P}+ \lambda_{cst} L_{cst}
\end{equation}
where $\lambda_{N}$, $\lambda_{P}$, and $\lambda_{cst}$ are balancing hyper-parameters.

\noindent
\textbf{Anomaly Score on Testing data. }
At test time, following the existing approaches for VAD \cite{liu2018future, park2020learning, astrid2021synthetic}, we predict frame-level anomaly scores and calculate these scores by using the Peak Signal to Noise Ratio (PSNR). The PSNR between the predicted frame $\hat{Y}$ and its ground truth $Y$ is used to compute the anomaly score as follows:

\begin{equation}
PSNR(Y, \hat{Y})=10 \log _{10} \frac{\left[\max _{\hat{Y}}\right]^{2}}{\frac{1}{N} \sum_{i=0}^{N}\left(Y_{i}-\hat{Y}_{i}\right)^{2}}
\end{equation}
where $N$ is the total number of pixels in $\hat{Y}$.
Then, we normalize the PSNR value to the range of [0,1] by a min-max normalization and calculate the regular score:
\begin{equation}
S(t)=\frac{P S N R\left(Y_{t}, \hat{Y}_{t}\right)-\min _{t} PSNR\left(Y_{t}, \hat{Y}_{t}\right)}{\max _{t} PSNR\left(Y_{t}, \hat{Y}_{t}\right)-\min _{t} PSNR\left(Y_{t}, \hat{Y}_{t}\right)}
\end{equation}
where $t$ is the frame index.

\section{Experiments and Results}

\textbf{Implementation Details. }
To evaluate the performance of the proposed framework, We conduct experiments on three challenging VAD datasets, \emph{i.e.}, UCSD Ped2 \cite{li2013anomaly}, CUHK Avenue \cite{lu2013abnormal} and ShanghaiTech \cite{luo2017revisit}. We initialize network parameters with the pre-trained weights of SimMIM on ImageNet-1k \cite{xie2021simmim} and employ an AdamW optimizer with $\beta_1$ = 0.9, $\beta_2$ = 0.999, and \emph{cosine} learning rate scheduler. We set the initial learning rate to be $1e^{-4}$, weight decay as 0.05, and warm-up for ten epochs. The batch size and epoch number of Ped2, Avenue, and ShanghaiTech are set to (4, 60), (4, 40), and (4, 10), respectively. $\lambda_{N}$, $\lambda_{P}$, and $\lambda_{cst}$ are 1.0, 1.0 and 0.3, respectively. The input size of all video frames is set to 224 × 224, and the default masked patch size is 32 × 32. The random mask ratio for generating pseudo anomalies of Ped2, Avenue, and ShanghaiTech is set to 0.75, 0.25, and 0.75, respectively.

\noindent
\textbf{Evaluation Criteria. }
To measure the VAD performance, we apply the frame-level Area Under the Receiver Operation Characteristic (AUROC) by varying the threshold over anomaly score \cite{liu2018future, astrid2021synthetic, astrid2021learning}, which is the widely used popular metrics for VAD. Higher AUC values indicate better VAD performance.


\setlength{\tabcolsep}{4pt}
\begin{table}
\begin{center}
\caption{
AUROC (\%) comparison between the proposed framework and state-of- the-art VAD methods on three public benchmarks.
}
\label{table:results}
\begin{tabular}{c|ccc}

    \hline
    Methods            & Ped2 & Avenue & SHTech \\ \hline
    ConvAE \cite{hasan2016learning}            & 90.0        & 70.2        & -            \\
    ConvLSTM-AE \cite{luo2017remembering}       & 88.1      & 77.0          & -            \\
    MemAE \cite{gong2019memorizing}             & 94.1      & 83.3        & 71.2         \\
    MNAD \cite{park2020learning}            & 90.2        & 82.8        & 69.8         \\
    LNRA \cite{astrid2021learning} & 94.77     & 84.91       & 72.46        \\
    StackRNN \cite{luo2017revisit}          & 92.2      & 81.7        & 68.0 \\
    Frame-Pred \cite{liu2018future}        & 95.4      & 85.1        & 72.8         \\
    AMC \cite{nguyen2019anomaly}          & 96.2      & 86.9        & -         \\
    AMMC-Net \cite{cai2021appearance}          & \textbf{96.6}      & 86.6        & 73.7         \\ \hline
    Ours            & 95.5      & \textbf{87.63}       & \textbf{76.57}        \\ \hline

    \end{tabular}
\end{center}
\end{table}

\noindent
\textbf{Comparison with Existing Methods. }
As shown in Table 1, we compare our framework with several state-of-the-art methods on three public VAD datasets. As can be observed, the proposed approach achieves a comparable performance. Specifically, compared to reconstruction based methods, our method achieves better performance on all datasets. Compared to hybrid methods (\emph{e.g.} AMC \cite{nguyen2019anomaly}, AMMC-Net \cite{cai2021appearance}), our method demonstrates a comparable performance. However, these hybrid methods are designed with complex architectures. In contrast, our method supports simple end-to-end training and applies the universal AEs architecture. In particular, our model gets very significant improvement on the very large and challenging ShanghaiTech datasets. \\
\indent
To demonstrate our advantages compared to similar pseudo anomaly based methods, We also rigorously conduct a comparison between our framework and LNRA\cite{astrid2021learning} with the same ViT architecture, considering that we use a different network architecture. The results are shown in Table \ref{table:aa} and Table \ref{table:at}. It demonstrates that the proposed method not only achieves a better VAD performance but also significantly reduces the time cost for training.

\noindent
\textbf{Ablation Studies. }
We perform the ablation study to investigate the impact of different components in our framework: pseudo anomalies synthesized by random mask and the proposed normalcy consistency training strategy (NCT). Firstly, we use a universal AE. Secondly, following the previous training strategy \cite{astrid2021synthetic, astrid2021learning}, we take pseudo anomaly examples as input with a small probability. For NCT, we input both normal and corresponding pseudo-anomaly samples at the same time. Then we minimize the bidirectional KL-divergence between their encoding features for the purpose of mining the consistent normalcy knowledge. The result is shown in Table \ref{table:ablation}. We can observe that the introduction of pseudo anomalies and NCT helped to get a significant improvement. We think that NCT can help the model to learn a better representation of normal regions despite there occurring anomalous regions in a frame. So the model learns vivid reconstruction boundaries between normal and abnormal data, which leads to better performance for VAD.

\begin{table}
\begin{center}
\caption{
AUROC (\%) results of comparison between our framework and LNRA with the same ViT architecture.
}
\label{table:aa}
\begin{tabular}{c|ccc}
\hline
           & Ped2 & Avenue & SHTech \\ \hline
 LNRA\cite{astrid2021learning}            & 94.12      & 85.94           & 74.42            \\
Ours            & 95.5        & 87.63        & 76.57            \\
\hline
\end{tabular}
\end{center}
\end{table}
\setlength{\tabcolsep}{1.3pt}

\setlength{\tabcolsep}{8pt}
\begin{table}
\begin{center}
\caption{
Training time every epoch (\MakeLowercase{hours}) of comparison between our framework and LNRA with the same ViT architecture.
}
\label{table:at}
\begin{tabular}{c|ccc}
\hline
           & Ped2 & Avenue & SHTech \\ \hline
 LNRA\cite{astrid2021learning}            &   0.736   &    4.6        & 25.2            \\
Ours            & 0.034        & 0.25        & 2.9            \\
\hline
\end{tabular}
\end{center}
\end{table}

\begin{table}
\centering
\caption{
Ablation studies of each component in our framework on the ShanghaiTech dataset.
}
\label{table:ablation}
\begin{threeparttable}
    \begin{tabular}{c|cc|c}
    \hline
             Index   & PASRM & NCT & AUC (\%)\\ \hline
      $\mathcal{A}$            & \XSolidBrush      & \XSolidBrush       & 73.2      \\
     $\mathcal{B}$            & \CheckmarkBold      & \XSolidBrush            & 75.46            \\
     $\mathcal{C}$   &  \CheckmarkBold  & \CheckmarkBold  & 76.57 \\
    \hline
    \end{tabular}
    \begin{tablenotes}
			\scriptsize
			\item PASRM = pseudo anomalies synthesized by random mask.
			\item NCT = normalcy consistency training strategy.
		\end{tablenotes}
\end{threeparttable}
\end{table}

\noindent
\textbf{Visualization. }
We give some abnormal visualization examples in three challenging datasets to visually show how our proposed method helps discriminate anomalies. In Figure 2, AEs means using the same AE architecture as our framework for future frame prediction, which as a comparison, can directly show the impact of our method. From left to right, we show the ground-truth frames, prediction results of AEs, prediction results of ours, difference maps of AEs, and difference maps of ours. Moreover, these difference maps are generated by calculating the squared error of each pixel between the ground-truth frame and the predicted frame, followed by a min-max normalization. As seen from the figure, our framework can generate a larger prediction error for the abnormal region, which results in better performance for VAD. Precisely, our model further distorts anomalous appearance patterns, such as the bicycle in the first row and third row of Figure 2. For irregular motion, our model prefers to use the normalcy representations learned from training to predict the expected movements. For example, for the bag thrown into the air in the second row, our model fails to predict it because it is not expected. However, the comparison overfits it and predicts wrong information. These demonstrate that our approach can help the model to learn superior anomaly discrimination capability.

\begin{figure}[!htbp]
\centering
\setlength{\belowcaptionskip}{-0.8cm}
\includegraphics[width=85mm]{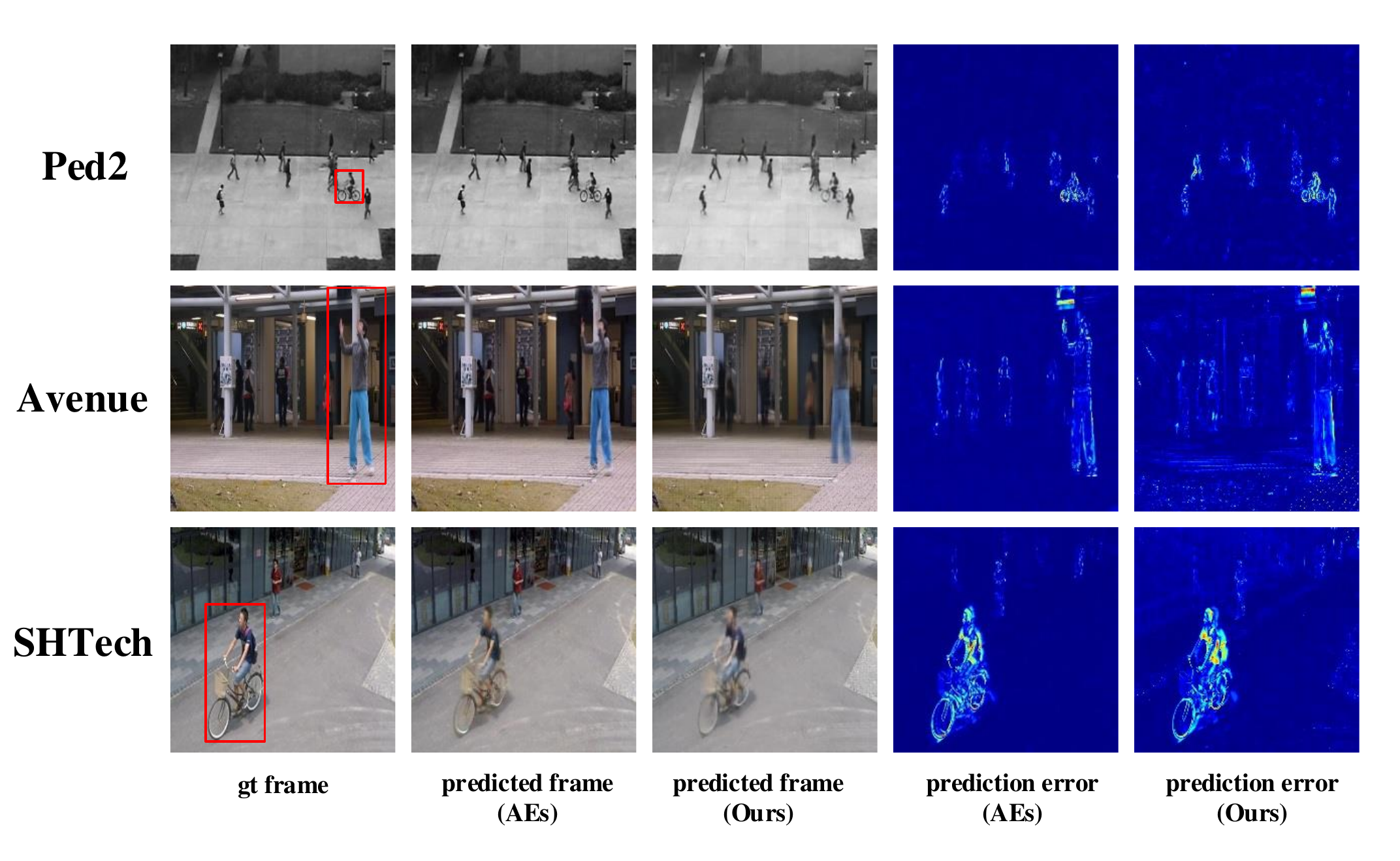}
\caption{Visualization examples of the target frame (gt frame), generated frame (predicted frame), and their difference (prediction error) in Ped2, Avenue, and ShanghaiTech datasets, respectively. The abnormal regions are marked with red boxes. Best viewed in color.
}
\label{fig:heatmap}
\end{figure}

\section{Conclusion}

In this paper, we propose a simple yet efficient framework based on masked autoencoder for unsupervised video anomaly detection. We first introduce the pseudo anomaly samples during training, which are synthesized from only normal data by embedding random mask tokens. Then, we propose a normalcy consistency training strategy to regular the representations from normal and corresponding pseudo anomaly data, which helps the model to adequately learn normalcy representations despite the presence of anomalous regions in the frame. The proposed approach can help to build more distinct reconstruction boundaries between normal and abnormal data. Extensive experiments on three challenging video anomaly detection datasets demonstrate the effectiveness of our proposed framework.

\vfill\pagebreak

\bibliographystyle{IEEEbib}
\bibliography{strings,refs}

\end{document}